# NAIVE: A METHOD FOR REPRESENTING UNCERTAINTY AND TEMPORAL RELATIONSHIPS IN AN AUTOMATED REASONER

Michael C. Higgins[1]
Hewlett-Packard Laboratories
1651 Page Mill Road (28C)
Palo Alto, CA 94304

**Abstract**

This paper describes NAIVE, a low-level knowledge representation language and inferencing process. NAIVE has been designed for reasoning about nondeterministic dynamic systems like those found in medicine. Knowledge is represented in a graph structure consisting of nodes, which correspond to the variables describing the system of interest, and arcs, which correspond to the procedures used to infer the value of a variable from the values of other variables. The value of a variable can be determined at an instant in time, over a time interval or for a series of times. Information about the value of a variable is expressed as a probability density function which quantifies the likelihood of each possible value. The inferencing process uses these probability density functions to propagate uncertainty. NAIVE has been used to develop medical knowledge bases including over 100 variables.

## 1 Introduction

The development of an automated reasoner for medical domains provides many challenging problems. A medical reasoning system must be able to manipulate a wide range of qualitative as well as quantitative information. General medical knowledge can be uncertain since current understanding of physiology and pathophysiology is incomplete. Patient-specific knowledge often is uncertain because timely observations of the data used to model the patient may not be available. The clinical observation process which collects these data can be inexact. Finally, patients are dynamic systems: what is true at one moment may not be true at a later time. Moreover, the pattern of change often is significant to understanding a patient's status.

The focus of this paper is the management of time and uncertainty in NAIVE [2], a knowledge representation language that has been designed for reasoning about nondeterministic dynamic systems like those found in medicine. NAIVE is based on a knowledge representation that is similar to that used in SYNTEL [1] and DEMOS [2]. Knowledge is expressed in terms of variables describing the system of interest, such as a patient. Some knowledge is imperative (e.g., the patient's *Birthday* is February 14, 1967). Other knowledge is procedural (e.g., the patient's *Age* can be determined by subtracting the patient's *Birthday* from the current date). The structure of the procedural knowledge can range from simple arithmetic or logical combinations to complicated high-level models.

The contribution of NAIVE is the addition of a robust structure for incorporating time into the knowledge base and inferencing process. The value of a variable can be determined at an instant in time, over a time interval, or for a series of time instants. The representation of time is incorporated in the expression of procedural knowledge so that knowledge bases can be developed which support reasoning about the dynamics of variables. For example, inferences can be made about the presence of trends. Inferences about trends, in turn, can be incorporated into inferences about other variables.

---

[1] The author wishes to acknowledge the contribution of Dr. James Lindauer to this work.

[2] The name "NAIVE" has been chosen to emphasize the inability of a functional representation system, like this one, to resolve contradictions in the knowledge it is given. In anthropomorphic terms, this system believes what it is told.

The goal of the following discussion is a description of the types of knowledge that can be encoded and manipulated with NAIVE. Section 2 describes the representation of uncertainty and Section 3 describes the inferencing process built on this uncertainty representation. Section 4 outlines the representation of procedural knowledge. The penultimate section discusses two implementation concerns.

## 2  Uncertainty Representation

Any attribute of the system of interest can be a *variable* in the knowledge base. The values for some attributes are the same for all time. These time invariant attributes are called *constants*. Since constants can be viewed as a subclass of variables, this discussion will focus on the latter.

The value of a variable can be determined for different types of time. Time can be an *instant* (e.g., the patient's *Weight* at *8:00 AM on Day 1* was 70.0 kg). Time can be an *interval* (e.g., the patient's *WeightChange* from *8:00 AM on Day 1 to 8:00 AM on Day 10* was *-2.0 kg*). Time can be a *series* (e.g., the serum samples collected at *8:00 AM, 8:30 AM, and 10:00 AM on Day 3* were used to determine the patient's *Glucose Tolerance*).

One of the defining properties of a variable is the *range* of values that it can assume. In NAIVE the elements in a range are collectively exhaustive and mutually exclusive. The notation $\Omega_X$ denotes the range for variable $X$. A variable can be classified according to the type of elements in its range. For example, the patient's *Sex* is a *categorical* constant with the range:

$$\Omega_{Sex} = \{female, male\}$$

The patient's *Glucose* level can be represented as an *ordinal* variable with the range:

$$\Omega_{Glucose} = \{hypoglycemia, normoglycemia, hyperglycemia\}$$

The patient's *Weight* can be represented as a *cardinal* variable with the range:

$$\Omega_{Weight} = \{x \mid 1\,kg \leq x \leq 300\,kg\}$$

The value of a variable can be the parameters in a function. For example, the *LinearTrend* in the patient's body weight can be represented as a function-valued variable with the range:

$$\Omega_{LinearTrend} = \{(\alpha, \beta) \mid -\infty < \alpha < \infty \text{ and } -\infty < \beta < \infty\}$$

where $\alpha$ and $\beta$ are the zeroth and first-order coefficients, respectively, for a linear function fitted to weight observations. The values for *LinearTrend* would be determined for time intervals.

In recent years a variety of quantitative and qualitative schemes have been proposed for the representation of uncertainty [3,4,5]. NAIVE uses traditional probability theory to represent uncertainty because of the minimal assumption set underlying this approach and the simple method of combination it provides [6]. The likelihood that variable $X$ has value $x$ at time $t$ is expressed as a probability density function $f_X(x,t)$ which has the property that for any set $\Gamma \subseteq \Omega_X$

$$Prob(X(t) \subseteq \Gamma) = \int_\Gamma f_X(x,t)\, dx$$

Given this uncertainty representation, an inexact statement about the value of a variable can be expressed as a uniform density function over a subset of the range. For example, the statement that the patient's *Age* at time $t$ is between 20 years and 30 years can be expressed as the density function

$$f_{Age}(x,t) = \begin{cases} \frac{1}{10\,years} & \text{if } 20\,years \leq x \leq 30\,years \\ 0 & \text{otherwise} \end{cases}$$

Conversely, complete certainty about the value of a variable is expressed as a Dirac delta function which is nonzero for exactly one element in the range.

## 3  Inferencing Process

The values of some variables are determined outside of the system. For example, values for *ReportedWeight* might be measured directly and reported as imperative knowledge. In NAIVE, a variable which has externally determined values is called a *datum*. A datum corresponds to a source node in the knowledge base graph. The values for all other variables are determined within the inferencing process by using procedural knowledge. An internally derived variable is called an *inference*. Specification of the procedural knowledge used to evaluate an inference is part of the definition of that variable. The classification of a variable as a datum or an inference depends on the design of the knowledge base; however, in a typical medical domain the patient's *Birthday* would be an example of a datum and the patient's *Age* would be an example of an inference.

The inferencing process is initiated by the evaluation of an inference. In general, an inference is evaluated by backward chaining to other variables in the knowledge base. This recursive process eventually terminates with the imperative knowledge that has been recorded for a datum. The next section discusses the expression of procedural knowledge in more detail; however, the following simplified examples illustrate the basic features of this important concept.

For a given time interval, the patient's total fluid *Intake* is the sum of *Oral* fluid intake and intravenous (*IV*) fluid intake. Symbolically,

$$Intake(t) = Oral(t) + IV(t)$$

where $t$ denotes the time interval of interest. From basic probability theory, the density function for *Intake* can be determined by the convolution integral

$$f_{Intake}(x,t) = \int_{\Omega_{Intake}} f_{Oral}(y,t) f_{IV|Oral}(x-y, y, t)\, dy$$

where $f_{IV|Oral}(x-y,y,t)$ is the probability density function for $IV(t) = x - y$ conditioned on $Oral(t) = y$. If the measurement of oral fluid intake is stochastically independent of the measurement of intravenous fluid intake then

$$f_{IV|Oral}(x-y,y,t) = f_{IV}(x-y,t)$$

in which case

$$f_{Intake}(x,t) = \int_{\Omega_{Intake}} f_{Oral}(y,t) f_{IV}(x-y,t)\, dy$$

In either case, any uncertainty in the value of *Oral* or *IV* would be reflected in the density function determined for *Intake*.

The representation of a single physiologic entity can require more than one variable in the knowledge base. The additional variables represent alternative methods for determining the value of the underlying physiologic entity. For example, in a simple knowledge base, the patient's body weight might be represented by a datum called *ReportedWeight*, an inference called *CurrentWeight* and a constant called *UnknownWeight*. The value for *CurrentWeight* is determined by the observations of *ReportedWeight* recorded within a given radius of the specified time. If an observation of the datum *ReportedWeight* is not available within the time interval, the value of *CurrentWeight* equals the density function for *UnknownWeight*. The density function for *UnknownWeight*, in turn, is uniformly distributed over the possible range for body weight. Thus, two evaluation methods have been combined for the same physiologic entity. Other strategies for combining procedural knowledge are discussed in the next section.

## 4 Representation of Procedural Knowledge

As noted earlier the definition of a variable that is classified as an inference includes the specification of a procedure for determining the value of that variable. The encoding of this procedural knowledge is fundamental to the structure of NAIVE. The types of procedural knowledge used by the inferencing process will depend on the domain; however, the following examples illustrate the functional forms used to encode medical knowledge bases.

### 4.1 Simple Variable Combinations

The convolution integral used to determine the probability density function for total fluid *Intake* in the previous section is one example of how procedural knowledge can be encoded for a cardinal variable. Similar expressions can be used to determine the probability density functions for variables that are derived by subtracting, multiplying or dividing variables [7, pages 316–318]. These four basic arithmetic operators can be used to build more complicated inferencing procedures such as polynomial combinations of variables. In a similar manner, higher level operators such as integration and differentiation can be implemented.

The value of a variable can be based on whether or not the value for another variable is contained in a set. For example, suppose that the value of the ordinal variable *GlucoseLevel* is *normoglycemia* if the value of the cardinal variable *SerumGlucose* is between 70 $mg/dl$ and 120 $mg/dl$. Then

$$f_{GlucoseLevel}(normoglycemia, t) = \int_{70}^{120} f_{SerumGlucose}(x)\, dx$$

Similar expressions can be used to determine the value of the density function for the other values in $\Omega_{GlucoseLevel}$. Inference procedures of this form can be used to transform the type of range for a variable, such as from cardinal to ordinal.

## 4.2 Time Interpolation and Extrapolation

Very few clinical variables are measured continuously. More typically, the observations of a variable are separated by time intervals of varying length. Therefore, the knowledge base must include interpolation and extrapolation procedures that can be used to estimate the value of a variable at times not coinciding with reported observations.

The example of *CurrentWeight* used to demonstrate the inferencing process in the previous section illustrates a simple version of time interpolation and extrapolation. The inferencing procedure for *CurrentWeight* uses the value reported for the related datum if an observation has been recorded at a time that is within a given radius of the time of interest. The length of the time radius is part of the procedural knowledge in the definition of *CurrentWeight* and can itself be an inference.

A more sophisticated inferencing method would be one which fits a parametric function to several observations reported at times adjacent to the time of interest. For example, *CurrentWeight* could be inferred by fitting a linear model to values recorded for *ReportedWeight*. The goodness-of-fit of the empirical model to the available data would determine the uncertainty in the inferred density. The resulting model can then be used to extrapolate the observed values. In the case of ordinal variables, interpolation or extrapolation can be based on qualitative models.

The inferencing procedure for some variables can be based on the causal behavior of the physiologic entity in question. For example, body weight changes as a result of the patient's metabolism and the difference between intake and output. Therefore, the value for *CurrentWeight* could be extrapolated from the most recent observations of *ReportedWeight* by determining the net balance in the patient's input, output and caloric requirements. Input, output and caloric requirements, in turn, can be determined from reported data and from inferenced values for other attributes in the knowledge base. Of course, the uncertainty in the density function inferred for *CurrentWeight* would reflect the uncertainty in the evaluation of these component variables.

Notice that the inferencing procedures described in this subsection change time from an instant into an interval. For example, the simple method of setting *CurrentWeight* equal to the nearest observation expands the instant at which a value has been recorded into an interval over which that value can be assumed to describe the patient. Inferencing procedures are described later in this section which convert time from intervals into instants.

## 4.3 Combining Inference Methods

The discussion in Section 3 noted that the knowledge base should accommodate alternative methods for determining the value of a physiologic entity. For example, the discussion in the previous subsection demonstrates that different procedures can be used to infer values for a single patient attribute, such a body weight. At another level, different medical tests can be used to measure the same entity. The following examples illustrate two general methods that can be used to combine several inferencing procedures.

A simple form of the first combining method that will be discussed was demonstrated by the example of *CurrentWeight*, *ReportedWeight*, and *UnknownWeight* in Section 3. The inferencing procedure described for that example included a primary procedure (e.g., the observation of *ReportedWeight* nearest the time of interest), a criterion for deciding if the primary procedure is valid (e.g., the time of the nearest reported value for *ReportedWeight* must be within a specified radius of the time of interest), and an alternative procedure (e.g., the density function for *UnknownWeight*).

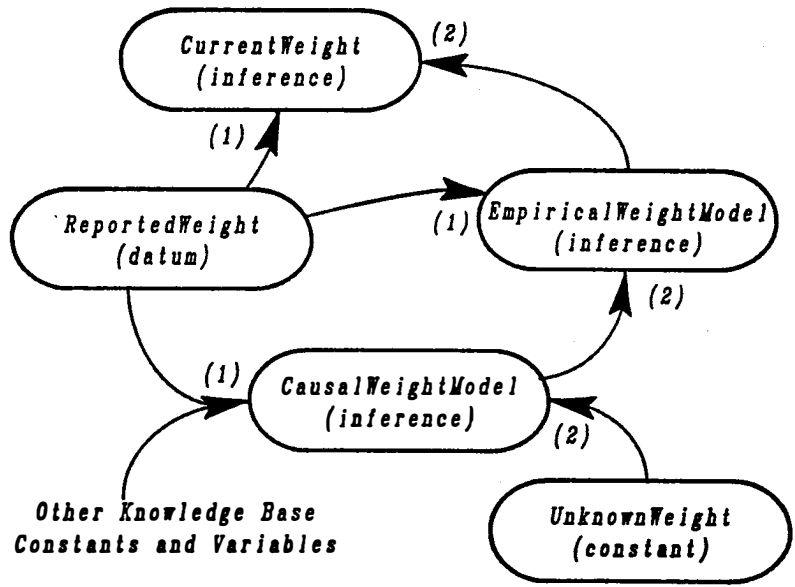

Figure 1: Portion of the knowledge base used to determine *CurrentWeight* from *ReportedWeight*, *CausalWeightModel*, *EmpiricalWeightModel* and *UnknownWeight*. The inferencing procedures are labelled (1) for primary and (2) for alternate.

This general approach can be used to combine the three interpolation and extrapolation methods discussed in Subsection 4.2 (see Figure 1). Let *CurrentWeight* be an inference which determines its values by using the nearest recorded value for the datum *ReportedWeight*. If a value has not been recorded for *ReportedWeight* within a specified time interval, *CurrentWeight* determines its value by evaluating an inference called *EmpiricalWeightModel*. If possible, *EmpiricalWeightModel* determines the patient's weight by fitting a linear model to the 10 observations of *ReportedWeight* that are nearest to the time of interest. Otherwise *EmpiricalWeightModel* determines its value by using the value for *CausalWeightModel*. If possible, *CausalWeightModel* uses a causal model to determine the patient's weight. Otherwise, *CausalWeightModel* assumes that the patient's weight is described by the uniform probability density function stored as a constant called *UnknownWeight*.

The combining method described above is based on a ranking of the possible inferencing procedures. The probability density function for *CurrentWeight* is determined by the highest ranked procedure which satisfies its validity criterion. Bayes formula provides an alternative that can be used to combine inferencing procedures that are based on conditionally independent data. For example, suppose that *Test 1* and *Test 2* are different measurements of the patient's serum *Glucose* level. Bayes formula can be transformed into the following relationship if *Test 1* and *Test 2* are conditionally independent given the patient's actual serum glucose level:

$$f_{Glucose}(x,t) = \frac{f_{Test1}(x,t) f_{Test2}(x,t)}{\int_{\Omega_{Glucose}} f_{Test1}(x,t) f_{Test2}(x,t) \, dx}$$

This expression can be expanded to combine an arbitrary number of methods for determin-ing the same physiologic entity provided that the knowledge base assumes that they are conditionally independent.

## 4.4 Trends

The trend in a variable over a time interval can be inferred based on ordinal comparisons. For example, suppose that *WeightTrend* is an ordinal variable with the following range:

$$\Omega_{WeightTrend} = \{decreasing, stable, increasing\}$$

The probability density function for *WeightTrend* can then be inferred, for time $t$, by the nested integral:

$$f_{WeightTrend}(decreasing, t) = \int_{Min.Wt.}^{Max.Wt.} f_{Weight}(x, t-\epsilon) \int_{Min.Wt.}^{x} f_{\Delta Weight}(y, t+\epsilon, x, t-\epsilon) \, dy \, dx$$

where $f_{\Delta Weight}(y, t+\epsilon, x, t-\epsilon)$ is the probability density function for $Weight(t+\epsilon) = y$ conditioned on $Weight(t-\epsilon) = x$. In turn, the conditional density function $f_{\Delta Weight}(y, t+\epsilon, x, t-\epsilon)$ can be inferred by combining inferencing procedures in the manner illustrated in the previous subsection. For example, $f_{\Delta Weight}(y, t+\epsilon, x, t-\epsilon)$ can be determined by weight observations reported near times $t-\epsilon$ and $t+\epsilon$. If these observations are unavailable, $f_{\Delta Weight}(y, t+\epsilon, x, t-\epsilon)$ can be determined by a causal model of how body weight changes.

Notice that inferencing procedures for trends change time from an interval into an instant. For example, the procedures discussed for *WeightChange* infer a characteristic of the patient at an instant based on the behavior of body weight over a time interval.

## 5 Implementation Considerations

This section discusses two issues that have arisen in the implementation of NAIVE.

## 5.1 Caching Density Functions

The evaluation of an inference, in a probabilistic system like NAIVE, can be computationally expensive. Therefore, time efficiency favors the storage of a probability density function once it has been inferred. On the other hand, the validity of a density function can be changed by the reporting of additional data. Therefore, truth maintenance considerations favor deriving a probability density function each time it is used.

A forward chaining process can be used to implement a compromise between these two conflicting concerns. A variable caches the probability density functions that have been determined. These density functions are stored until a new observation is reported for a datum that is used in the inferencing procedure. The variables potentially affected by a new observation are identified by forward chaining from the node corresponding to the reported datum. The density functions are removed from the caches for the identified variables. Thus, an inferred density function is stored until it has become invalid because of a change in the imperative knowledge.

## 5.2 Contradictions

In a probabilistic system like NAIVE, a contradiction can be defined as the occurrence of an event which has probability zero. Given this definition, the knowledge base in NAIVE can be extended to assess the consistency of the reported data. For example, recalling the variables used to illustrate the discussion in Subsection 4.3, suppose that the datum *ReportedWeight* is paired with the model-based inference *EmpiricalWeightModel*. A reported value for *ReportedWeight* can be compared to the probability density inferred for *EmpiricalWeightModel*. The reported value is inconsistent with the existing imperative and procedural knowledge if the inferred density function assigns a probability of zero to the observation.

# 6 Conclusion

The functional representation languages like SYNTEL and DEMOS provide a robust method for managing uncertainty because these languages support the direct encoding of the tools that have been developed in the traditional probability fields. These probabilistic tools include statistics, Bayesian analysis and stochastic processes. In a similar manner NAIVE demonstrates that functional representation languages also can incorporate the tools that have been developed for the analysis of dynamics.

# References

[1] Rene Reboh and Tore Risch, "SYNTEL: Knowledge Programming Using Functional Representations", *Proceedings of AAAI-86: Philadelphia, PA*, American Association for Artificial Intelligence, pp. 1003 – 1007, 1986.

[2] Max Henrion and Nina Wishbow, *Demos User's Manual. Version Three*, Department of Engineering and Public Policy, Carnegie-Mellon University, Pittsburg, 1987.

[3] Edward H. Shortliffe and Bruce G. Buchanan, "A Model of Inexact Reasoning in Medicine", In *Rule-Based Expert Systems*, Addison-Wesley, pp 233 – 262, 1984.

[4] Lotfi Zadeh, "The Role of Fuzzy Logic in the Management of Uncertainty in Expert Systems", *Fuzzy Sets and Systems*, 11, pp 199 – 227, 1983.

[5] Glenn Shafer, *A Mathematical Theory of Evidence*, Princeton University Press, Princeton, NJ, 1976.

[6] Eric Horvitz, David Heckerman, Curt Langlotz, "A Framework for Comparing Alternative Formalisms for Plausible Reasoning", *Proceedings of AAAI-86: Philadelphia, PA.*, American Association for Artificial Intelligence, 1986.

[7] Emanuel Parzen, *Modern Probability Theory and Its Applications*, John Wiley, New York, 1960.